\documentclass[10pt,twocolumn,letterpaper]{article}

\usepackage[pagenumbers]{cvpr} 

\usepackage{graphicx}
\usepackage{amsmath}
\usepackage{amssymb}
\usepackage{booktabs}
\usepackage{comment}
\usepackage{adjustbox}
\usepackage{url}

\usepackage {colortbl,array,xcolor}

\newcommand{\figdir}{figure}

\usepackage[pagebackref,breaklinks,colorlinks]{hyperref}

\usepackage[capitalize]{cleveref}
\crefname{section}{Sec.}{Secs.}
\Crefname{section}{Section}{Sections}
\Crefname{table}{Table}{Tables}
\crefname{table}{Tab.}{Tabs.}

\def\cvprPaperID{4105} 
\def\confName{CVPR}
\def\confYear{2022}

\begin{document}

\title{Deep Depth from Focal Stack with Defocus Model for Camera-Setting Invariance}

\author{Yuki Fujimura$^1$ 
\and 
Masaaki Iiyama$^2$ 
\and 
Takuya Funatomi$^1$
\and 
Yasuhiro Mukaigawa$^1$ \\
$^1$Nara Institute of Science and Technology, Japan \quad $^2$Shiga University, Japan\\
{\tt\small fujimura.yuki@is.naist.jp} \quad {\tt\small iiyama@iiyama-lab.org} \\
{\tt\small \{funatomi,mukaigawa\}@is.naist.jp}
}
\maketitle

\begin{abstract}
We propose a learning-based depth from focus/defocus (DFF), which takes a focal stack as input for estimating scene depth. Defocus blur is a useful cue for depth estimation. However, the size of the blur depends on not only scene depth but also camera settings such as focus distance, focal length, and f-number. Current learning-based methods without any defocus models cannot estimate a correct depth map if camera settings are different at training and test times. Our method takes a plane sweep volume as input for the constraint between scene depth, defocus images, and camera settings, and this intermediate representation enables depth estimation with different camera settings at training and test times. This camera-setting invariance can enhance the applicability of learning-based DFF methods. The experimental results also indicate that our method is robust against a synthetic-to-real domain gap, and exhibits state-of-the-art performance.
\end{abstract}

\section{Introduction}
In computer vision, depth estimation from two-dimensional (2D) images is an important task and used for many applications such as VR, AR, or autonomous driving.
Defocus blur is a useful cue for such depth estimation because the size of the blur depends on scene depth.
Depth from focus/defocus (DFF) takes defocus images as input for depth estimation.
Typical inputs for DFF are stacked images, \ie, {\it focal stack}, each of which is captured with a different focus distance.

DFF methods are roughly divided into two categories, model-based and learning-based.
Model-based methods use a thin-lens model for modeling defocus blurs \cite{Suwajanakorn15,Kim16,Tang17} or define focus measures \cite{Pertuz13, Jaeheung17} to estimate scene depth.
One of the drawbacks of such methods is difficulty in estimating scene depth with texture-less surfaces.
Learning-based methods have been proposed to tackle the above drawback \cite{Hazirbas18,Maximov20,Wang21}.
For example, Hazirbas \etal \cite{Hazirbas18} proposed a convolutional neural network (CNN) taking a focal stack as input without any explicit defocus models.
This is an end-to-end method that allows efficient depth estimation.
It also enables the depth estimation of texture-less surfaces with learned semantic cues.

\begin{figure}[tb]
\centering
  \begin{minipage}[t]{0.48\linewidth}
    \centering
    \includegraphics[width=0.8\textwidth]{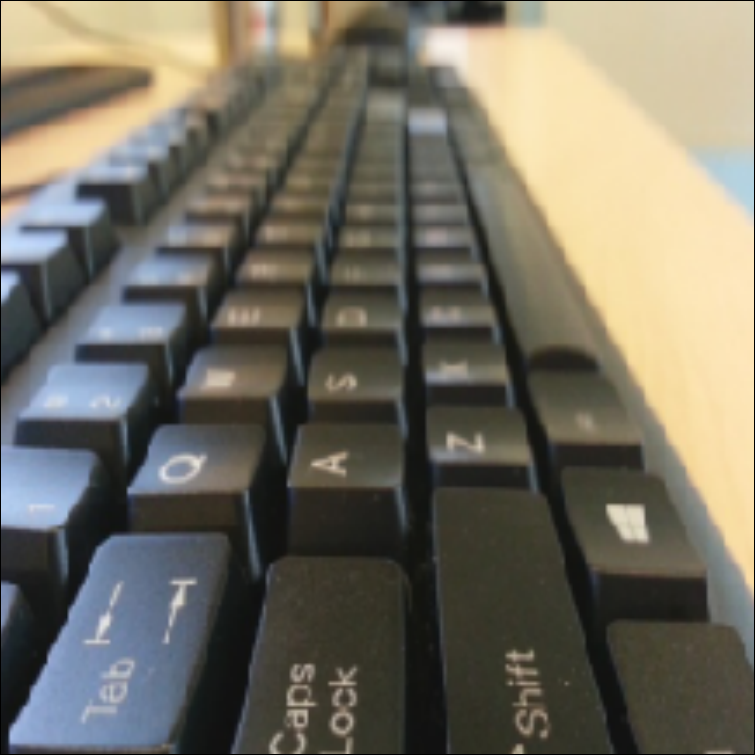}
    \subcaption{RGB}
  \end{minipage}
  \begin{minipage}[t]{0.48\linewidth}
    \centering
    \includegraphics[width=0.8\textwidth]{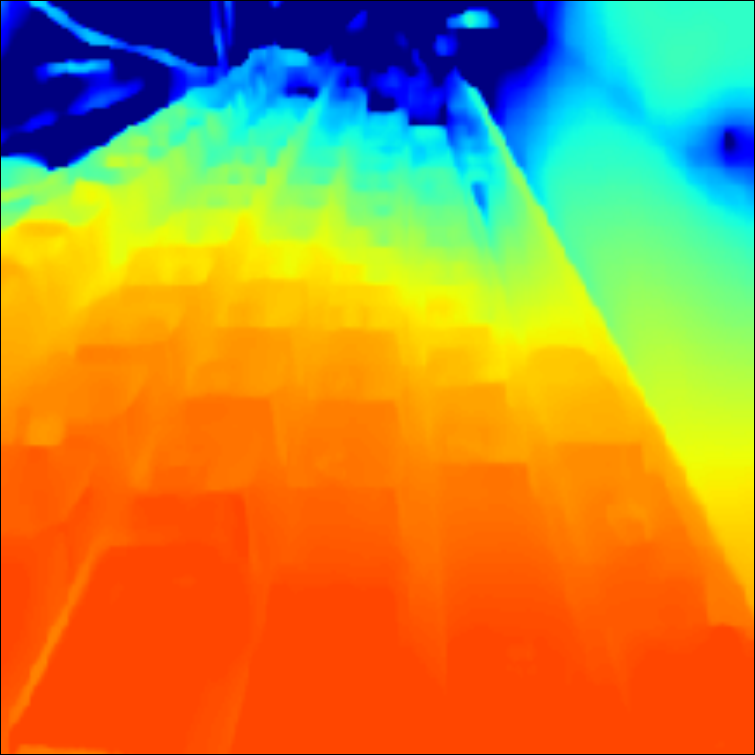}
    \subcaption{Suwajanakorn \etal \cite{Suwajanakorn15}}
  \end{minipage}\\
  \begin{minipage}[t]{0.48\linewidth}
    \centering
    \includegraphics[width=0.8\textwidth]{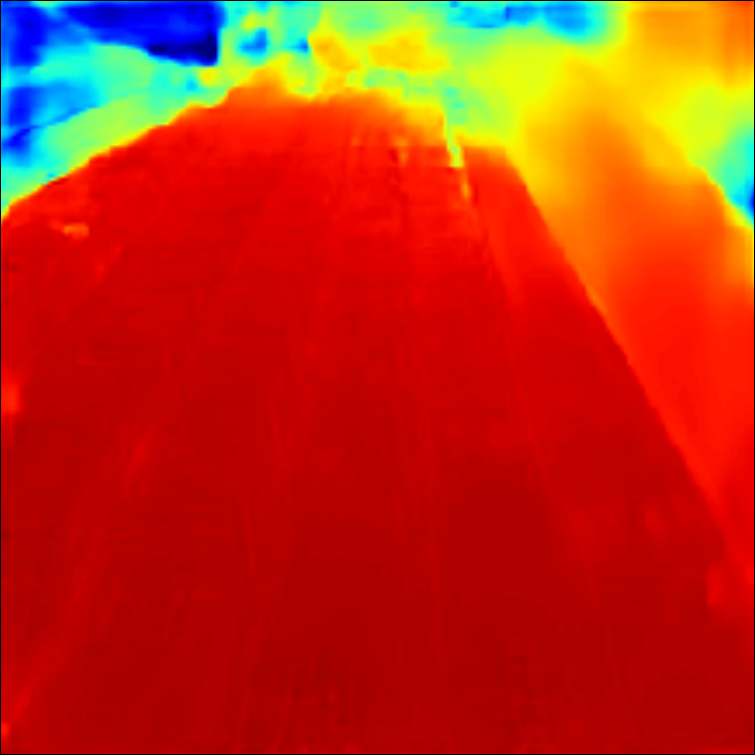}
    \subcaption{DefocusNet \cite{Maximov20}}
  \end{minipage}
  \begin{minipage}[t]{0.48\linewidth}
    \centering
    \includegraphics[width=0.8\textwidth]{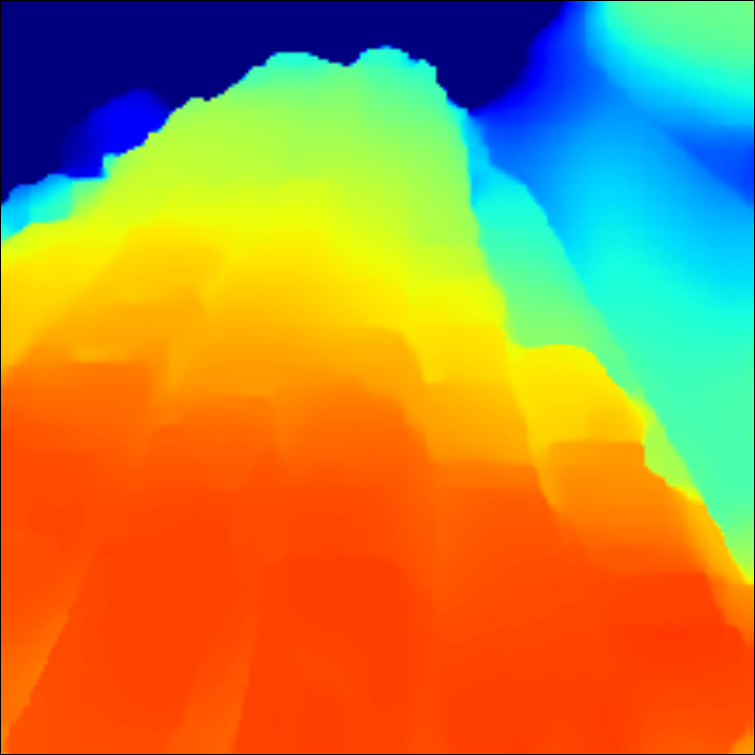}
    \subcaption{Ours}
  \end{minipage}
  \caption{(a) One of input images in focal stack, (b) output depth of \cite{Suwajanakorn15}, (c) output depth of DefocusNet \cite{Maximov20}, and (d) our result. Our model and DefocusNet were trained on a dataset with camera settings that differed from those of the test data. Our method with camera-setting invariance can estimate correct depth map.}
  \label{fig:cover_art}
\end{figure}

General learning-based methods often have limited generalization due to a domain gap between training and test data.
Learning-based DFF methods suffer from the difference of capture settings of a camera at training and test times.
The amount of a defocus blur depends on not only scene depth but also camera settings such as focus distance, focal length, and f-number. 
Different depths and camera settings can generate defocus images with the same appearance; thus this difference cannot be compensated with often used domain adaptation method such as neural style transfer \cite{Li17,Zhu17}.
If camera settings are different at training and test times, the estimated depth has some ambiguity, which is similar to the scale-ambiguity in monocular depth estimation \cite{Hu19}.
Current learning-based DFF methods \cite{Hazirbas18,Maximov20,Wang21} do not take into account the latent defocus model, thus the estimated depth is not correct if the camera settings at test time differ from those at training time, as shown in Fig. \ref{fig:cover_art}(c).
On the other hand, this problem does not matter for model-based methods with explicit defocus models under given camera settings.

We propose learning-based DFF with a lens defocus model.
Our method is inspired by recent learning-based multi-view stereo (MVS) \cite{Wang18}, where a cost volume is constructed on the basis of a plane sweep volume \cite{Collins96}.
The proposed method also constructs a cost volume, which is passed through a CNN to estimate scene depth.
Each defocus image in a focal stack is deblurred at each sampled depth in the plane sweep volume, then the consistency is evaluated between deblurred images.
We found that scene depth is effectively learned from the cost volume in DFF.
Our method has several advantages over the other learning-based  methods directly taking a focal stack as input without an explicit defocus model \cite{Hazirbas18,Maximov20,Wang21}.
First, output depth satisfies the defocus model because the cost volume imposes an explicit constraint among the scene depth, defocus images, and camera settings.
Second, the camera settings, such as focus distances and f-number are absorbed into the cost volume as intermediate representation.
This enables depth estimation with different camera settings at training and test times, as shown in Fig. \ref{fig:cover_art}(d).

The primary contributions of this paper are summarized as follows:
\begin{itemize}
  \item To the best of our knowledge, this is the first study to combine a learning framework and model-based DFF through a plane sweep volume.
  \item Our method with camera-setting invariance can be applied to datasets with different camera settings at training and test times, which improves the applicability of learning-based DFF methods.
  \item Similar to the previous learning-based method \cite{Maximov20}, our method is also robust against a synthetic-to-real domain gap and achieves state-of-the-art performance.
\end{itemize}

\section{Related work}
\paragraph{Depth from focus/defocus}
Depth from focus/defocus (DFF) estimates scene depth from focus or defocus cues in captured images and is a major task in computer vision.
In general, depth from focus takes many images captured with different focus distances and determines scene depth from an image with the best focus. On the other hand, depth from defocus aims to estimate scene depth from a small number of images, which do not necessarily need to include focused images \cite{Xiong93}.
Recently, depth estimation from a focal stack implicitly uses both focus and defocus cues; thus, we use unified terminology, {\it depth from focus/defocus}.

Traditional DFF methods propose focus measures to evaluate the amount of a defocus blur \cite{Zhuo11, Pertuz13, Moeller15, Jaeheung17}.
If we have a focal stack as input, we can simply refer to the image with noticeable edges and its focus distance.
Other methods formulate the amount of defocus blur with a lens defocus model and solve an optimization problem to obtain a depth map together with an all-in-focus image \cite{Suwajanakorn15,Kim16}.
We refer to these methods as model-based methods.
One of the drawbacks of such methods is difficulty in estimating scene depth with texture-less surfaces.
Learning-based methods have been proposed to tackle these issues \cite{Hazirbas18,Maximov20,Wang21}. 
These methods enable depth estimation at texture-less surfaces and the depth estimation is achieved efficiently in an end-to-end manner.
Other learning-based methods leveraged defocus cues as additional information \cite{Anwar17,Carvalho18} or supervision \cite{Srinivasan18, Gur19} for monocular depth estimation.

However, current learning-based DFF methods, which directly take a focal stack as input, do not take into account the latent defocus model \cite{Hazirbas18,Maximov20,Wang21}.
For example, Hazirbas \etal \cite{Hazirbas18} proposed a CNN that directly takes a focal stack as input. Maximov \etal \cite{Maximov20}  and Wang \etal \cite{Wang21} simply used focus distances as intermediate inputs of neural networks.
These methods require the same camera settings at training and test times to obtain a correct depth map due to the lack of explicit defocus models.
This characteristic reduces the applicability of learning-based DFF methods.
On the other hand, our method is a combination of model-based and learning-based methods through a cost volume, which is computed with a lens defocus model, allowing depth estimation with camera-setting invariance.

\paragraph{Learning from cost volume}
Learning from a cost volume is efficient in many applications.
A cost volume is constructed by sampling solution space and evaluating costs at each sampled point.
Examples of learning-based methods with a cost volume are optical flow \cite{Ilg17,Sun18} and disparity \cite{Mayer16,Kendall17} estimation.
Learning-based MVS methods \cite{Wang18, Yao18, Long21, Duzceker21} are also major examples, where a cost volume is constructed on the basis of a plane sweep volume \cite{Collins96}.
Our method also constructs a plane sweep volume and evaluates consistency between defocus images in an input focal stack.
We found that learning from a cost volume is also efficient for learning-based DFF.

\begin{figure*}[tb]
  \centering
  \includegraphics[width=0.9\textwidth]{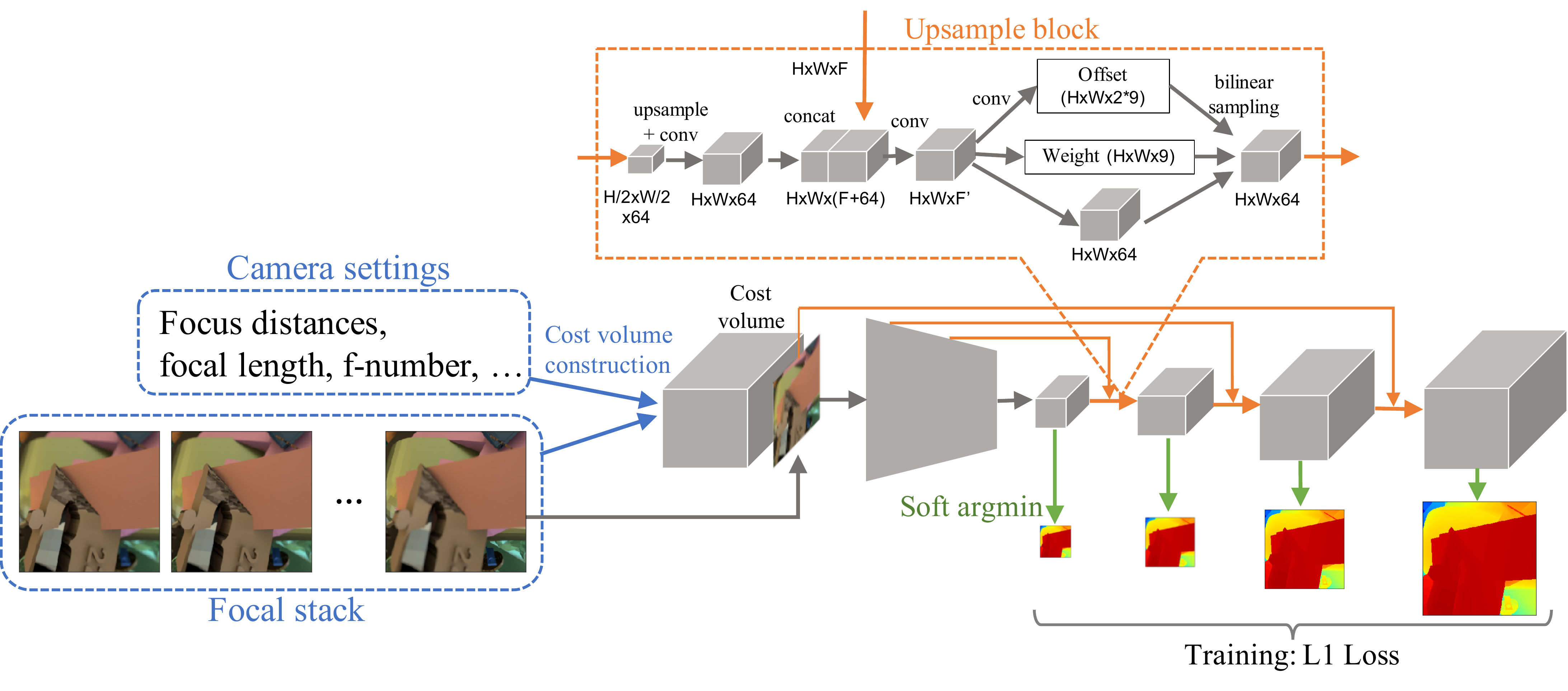}
  \caption{Overview of our method. Our method takes focal stack and camera settings as input then constructs cost volume as intermediate representation, which absorbs differences in camera settings. CNN takes this cost volume together with additional image as input then estimates refined cost volume in coarse-to-fine manner. Depth maps are computed by applying soft argmin operator at each resolution. Each upsample block has adaptive cost aggregation module.}
  \label{fig:overview}
\end{figure*}

\section{Deep depth from focal stack}
Our method combines a learning framework and model-based DFF through a cost volume for depth estimation with camera-setting invariance.
We first give an overview of the proposed method then describe the lens defocus model and ambiguity of estimated depth in DFF, followed by details of cost volume construction.
This cost volume as intermediate representation enables depth estimation with different camera settings at training and test times.
The network architecture and loss function are also discussed at the end of this section.

\subsection{Overview} \label{sec:overview}
Figure \ref{fig:overview} shows an overview of the proposed method.
Our method is inspired by recent learning-based MVS \cite{Wang18}, where a cost volume is constructed on the basis of a plane sweep volume \cite{Collins96}.
Our cost volume is constructed from an input focal stack by evaluating deblurred images at each depth hypothesis.
This intermediate representation absorbs the difference in camera settings.
The computed cost volume and an additional defocus image are passed through a CNN with an encoder-decoder architecture.
At the decoder part, the cost volume is gradually upsampled for coarse-to-fine estimation.
Output depth maps are obtained by applying a differentiable soft argmin operator \cite{Kendall17} to intermediate refined cost volumes.
Each upsample block includes a cost aggregation module for learning local structures adaptively.

\begin{figure}[tb]
  \centering
  \includegraphics[width=0.45\textwidth]{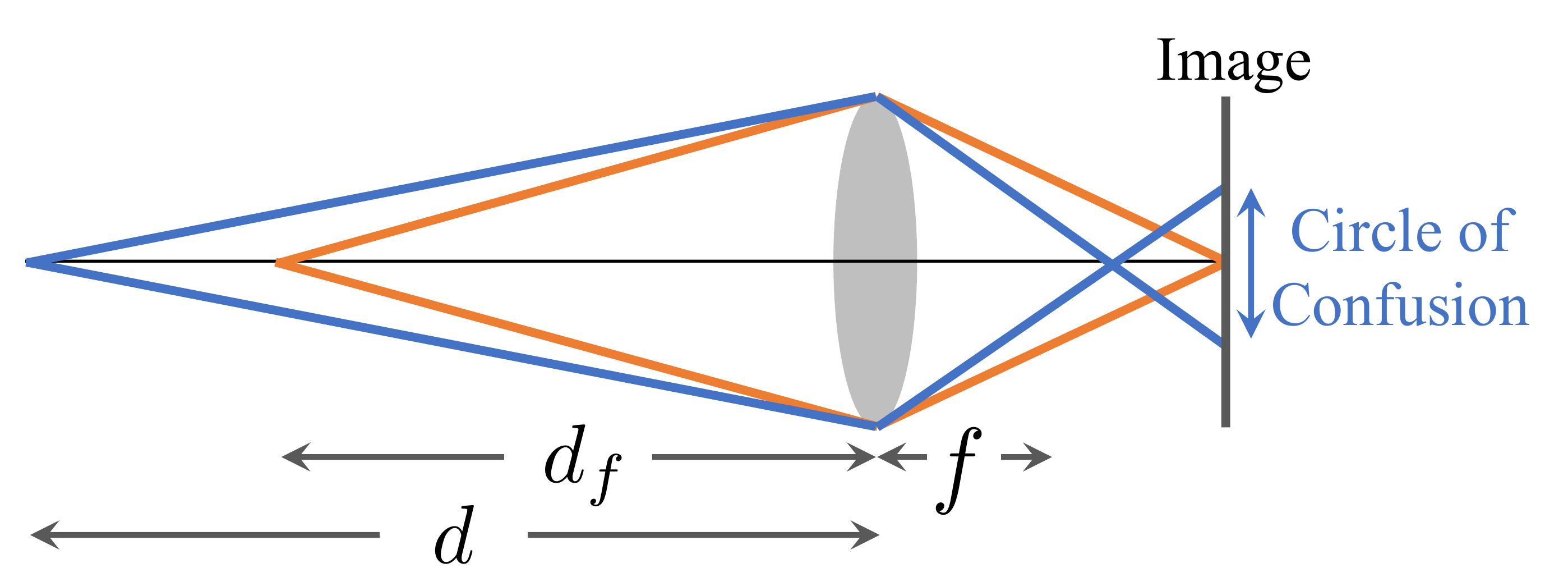}
  \caption{Circle of confusion (CoC) that corresponds to size of defocus blur}
  \label{fig:coc}
\end{figure}

\subsection{Lens defocus model} \label{sec:defocus_model}
Our cost volume construction is based on a lens defocus model, with which the size of a defocus blur is formulated as a circle of confusion (CoC) \cite{Zhuo11}, as shown in Fig. \ref{fig:coc}.
Let $d$ and $d_f$ be the scene depth and focus distance of a camera, respectively.
CoC can be computed as

\begin{equation}
c = b\frac{ \|d - d_f\| }{d} \frac{f^2}{N(d_f - f)},
\label{eq:defocus_model}
\end{equation}
where $f$ is the focal length of the lens and $N$ is the f-number.
$b$ [px/m] converts the unit of the CoC from [m] to [px].
When $d$ is equal to $d_f$, the light rays from the scene point converge on the image plane; otherwise, defocus blur results as the size of the diameter of the CoC.
The blurred image can be computed as a convolution of an all-in-focus image with the point spread function (PSF), the kernel size of which corresponds to the size of the CoC.

The CoC can be computed from the scene depth $d$ and the camera settings $f$, $d_f$, $N$, and $b$ in Eq.~(\ref{eq:defocus_model}).
Note that these parameters can easily be extracted from EXIF properties~\cite{Maximov20} or calibrated beforehand~\cite{Tang17}, and the state-of-the-art methods assume these parameters are known~\cite{Maximov20, Wang21}; thus, this paper also follows the same assumption.
Our method realizes depth estimation with camera-setting invariance about these parameters,
and this improves the applicability of learning-based DFF methods because our method with camera-setting invariance can be applied to datasets with different camera settings at training and test times.

Now, we discuss two ambiguities in DFF due to the camera settings.
The first one is scale-ambiguity.
From Eq. (\ref{eq:defocus_model}), the following relationship holds:
\begin{align}
c &= b\frac{ \|d - d_f\| }{d} \frac{f^2}{N(d_f - f)}&\nonumber\\
&= b^{-*}\frac{ \|d^* - d^*_f\| }{d^*} \frac{f^{*2}}{N(d^*_f - f^*)},&
\end{align}
where $(\cdot)^* = (\cdot)\sigma, (\cdot)^{-*} = (\cdot)/\sigma, \, \forall\sigma \in \mathbb{R}$.
This means scaled camera settings and depth give the same CoC as that of the original ones.

The other ambiguity is affine-ambiguity.
From Eq. (\ref{eq:defocus_model}), we can obtain
\begin{align}
c &= b\frac{f^2}{N(d_f-f)} \left\| 1 - \frac{d_f}{d} \right\|&\nonumber\\
&= A(f, d_f, N) + \frac{B(f, d_f, N)}{d},&
\end{align}
where $A(f, d_f, N)$ and $B(f, d_f, N)$ are constants.
Thus, different camera settings and inverse depths can give the same CoC as follows:
\begin{align}
c &= A(f, d_f, N) + \frac{B(f, d_f, N)}{d}& \nonumber \\
&= A(f', d'_f, N') + \frac{B(f', d'_f, N')}{d'}.&
\end{align}
This means the estimated inverse depth has affine-ambiguity (Similar discussion can be found in the previous study~\cite{Garg19}).
In the experiments, we evaluate the proposed method with respect to the scale-ambiguity in the depth space and the affine-ambiguity in the inverse depth space.

\subsection{Cost volume} \label{sec:cost_volume}
The proposed method computes a cost volume from the focal stack for the input of a CNN to impose a constraint between the defocus images and scene depth.
This has several advantages over current learning-based methods that directly takes a focal stack as input \cite{Hazirbas18,Maximov20,Wang21}.
First, output depth satisfies the lens defocus model because the cost volume imposes an explicit constraint between the defocus images and scene depth.
Second, the camera settings are absorbed into the cost volume.
This enables inference with camera settings that differ from those at training, and even in this case, the output depth satisfies the lens defocus model without any ambiguities. 

\begin{figure}[tb]
  \centering
  \includegraphics[width=0.5\textwidth]{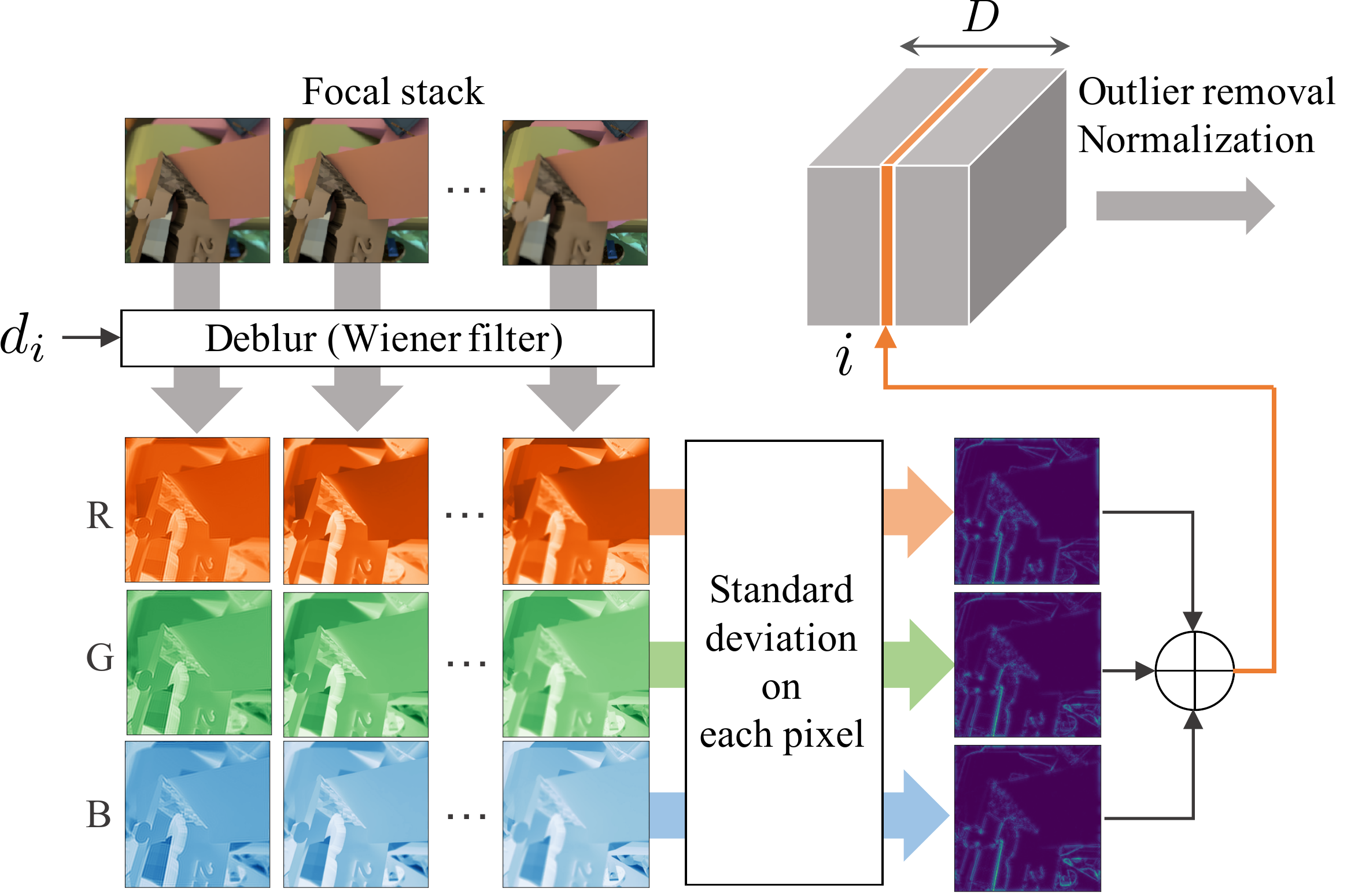}
  \caption{Cost volume construction. We first sweep fronto-parallel plane in camera coordinate system. At each swept plane, input image in focal stack is deblurred with Wiener–Hunt deconvolution \cite{Francois10} on each color channel. Standard deviation is applied for computing cost, which is followed by outlier removal and normalization.}
  \label{fig:cost_volume}
\end{figure}

Figure \ref{fig:cost_volume} shows a diagram of our cost volume construction.
We first sample the 3D space in the camera coordinate system by sweeping a fronto-parallel plane.
To evaluate each depth hypothesis, we deblur each image in the input focal stack.
Let the cost volume be $C:\{1,\cdots,W\}\times \{1,\cdots,H\}\times \{ 1,\cdots,D\} \rightarrow \mathbb{R}$, and the focal stack be 
$\{ I_{d_i}\}_{i=1}^{F}$, where $I_{d_i}$ is a captured image with focus distance $d_i$. 
Each element of the cost volume $C$ is computed as follows:
\begin{align}
C(u,v,d) &= \sum_{ch \in \{r,g,b\}}\rho\Bigl(\tilde{I}_{d_1}^{ch}(u,v),\cdots,\tilde{I}_{d_F}^{ch}(u,v) \Bigr),& \label{eq:cost_computation} \\
\tilde{I}_{d_i}^{ch} &= k(d,d_i) *^{-1} I_{d_i}^{ch},& 
\end{align}
where $k(d,d_i)$ is a blur kernel, the size of which is defined by Eq. (\ref{eq:defocus_model}) with the scene depth $d$ and focus distance $d_i$.
We used a disk-shaped PSF \cite{Watanabe98,Shi15}, while any types of PSFs can be used at training and test time.
The operator $*^{-1}$ indicates a deblurring process applied to each color channel of the input image.
We used Wiener–Hunt deconvolution \cite{Francois10} as this process.
The function $\rho$ evaluates the consistency between deblurred images.
We adopt a standard deviation for $\rho$, which allows an arbitrary number of inputs.
Note that a similar cost volume computation was proposed in model-based methods \cite{Suwajanakorn15,Kim16}.
However, these methods require an all-in-focus image, which leads to iterative optimization for the scene depth and all-in-focus image; thus these methods cannot be directly incorporated into sequential learning frameworks.

\begin{figure}[tb]
\centering
  \captionsetup{format=hang}
  \begin{minipage}[t]{0.4\linewidth}
    \centering
    \includegraphics[width=1.0\textwidth]{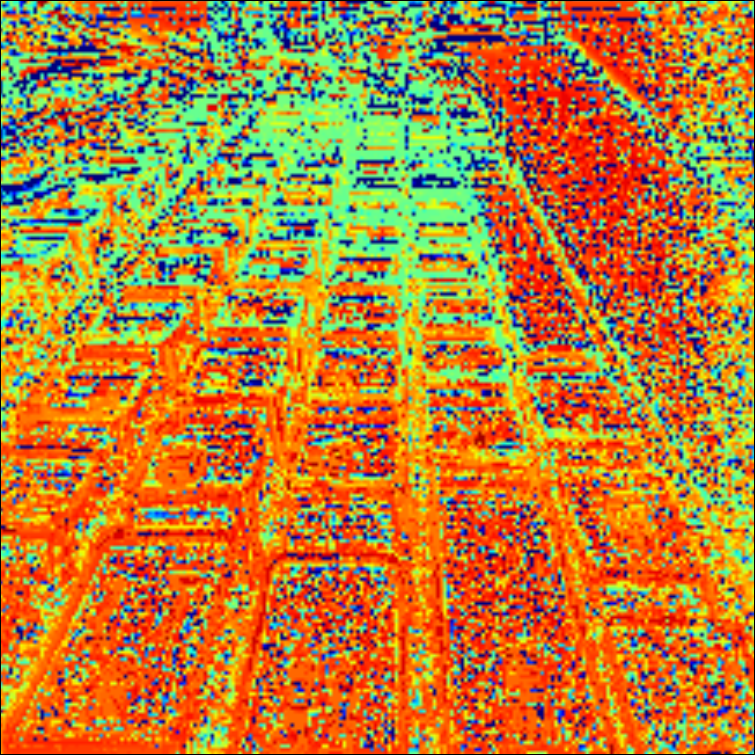}
    \subcaption{Without neighboring pixels}
  \end{minipage}
  \begin{minipage}[t]{0.4\linewidth}
    \centering
    \includegraphics[width=1.0\textwidth]{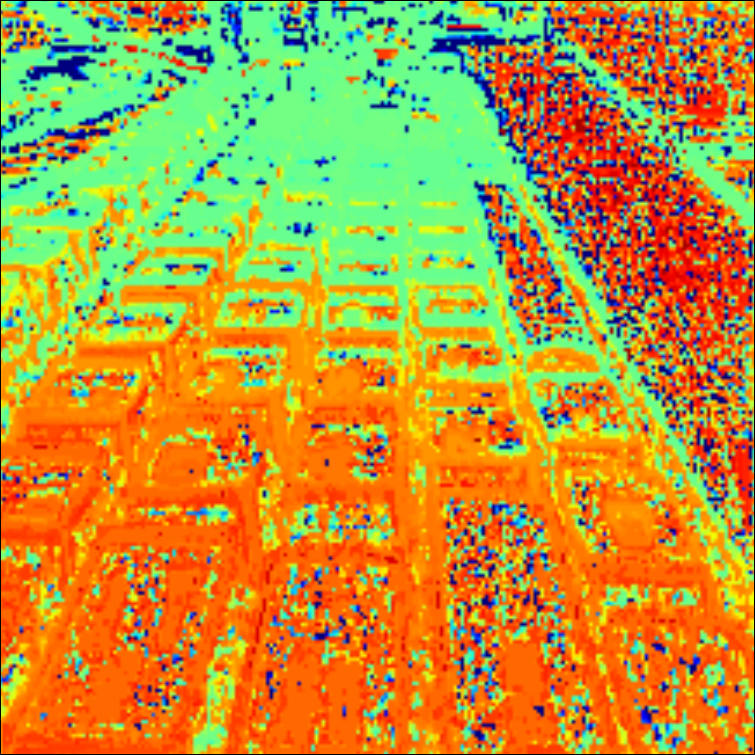}
    \subcaption{With neighboring pixels}
  \end{minipage}
  \captionsetup{format=plain}
  \caption{Estimated depth from the index of the minimum cost (a)~without and (b)~with the neighboring pixels.}
  \label{fig:non_local_cost}
\end{figure}

The process mentioned above is the essential part of our cost volume construction.
However, differing from a learning-based MVS method \cite{Wang18}, which is based on differentiable image warping, our cost volume construction requires careful design because the difference between images due to focus distances is smaller than that due to camera positions.
Thus, for robustness and learning stability, the standard deviation in Eq. (\ref{eq:cost_computation}) is computed considering neighboring pixels as follows:
\begin{flalign}
&\rho\Bigl(\tilde{I}_{d_1}^{ch}(u,v),\cdots,\tilde{I}_{d_F}^{ch}(u,v) \Bigr) & \nonumber \\
&= \sqrt{\frac{1}{F} \sum_{i=1}^F \sum_{(u',v') \in \mathcal{N}(u,v)} \gamma_{u',v'} (\tilde{I}_{d_i}^{ch}(u',v') - \mu(u',v'))^2 },&\label{eq:stddev}\\
&\mu(u,v) = \frac{1}{F} \sum_{i=1}^F \sum_{(u',v') \in \mathcal{N}(u,v)} \gamma_{u',v'} \tilde{I}_{d_i}^{ch}(u',v'),&
\end{flalign}
where $\mathcal{N}(u,v)$ is a set of neighboring pixels centered at $(u,v)$ and $\gamma_{u',v'}$ is a 2D spatial Gaussian weight.
Figure \ref{fig:non_local_cost} shows an example of the estimated depth only from the index of the minimum cost.
The neighboring information can reduce noise, especially for the real captured data.

\begin{figure}[tb]
\centering
  \captionsetup{format=hang}
  \begin{minipage}[t]{0.25\linewidth}
    \centering
    \includegraphics[width=1.0\textwidth]{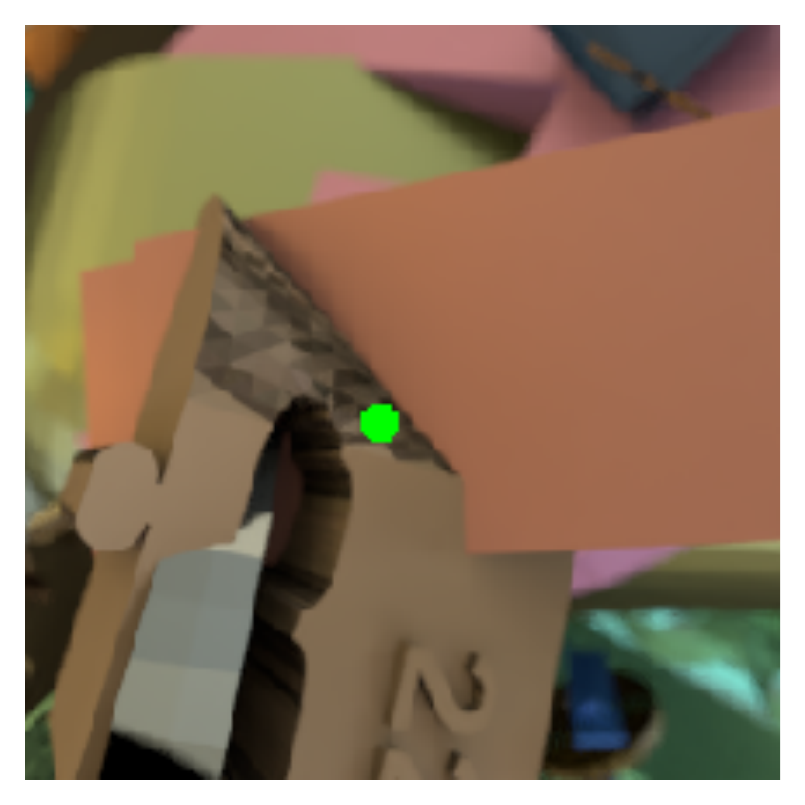}
    \subcaption{RGB}
  \end{minipage}
  \begin{minipage}[t]{0.35\linewidth}
    \centering
    \includegraphics[width=1.0\textwidth]{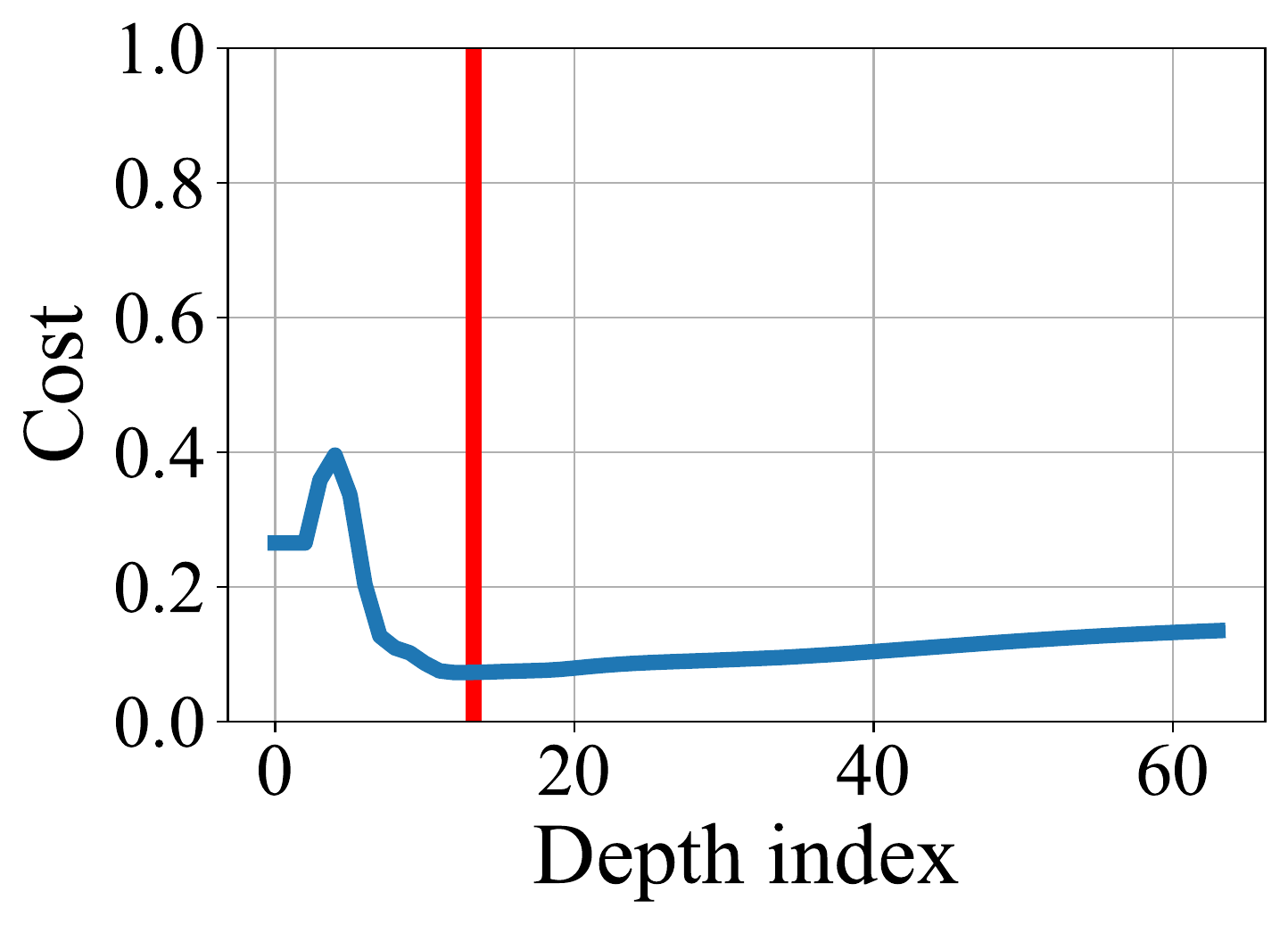}
    \subcaption{Initial cost}
  \end{minipage}
  \begin{minipage}[t]{0.35\linewidth}
    \centering
    \includegraphics[width=1.0\textwidth]{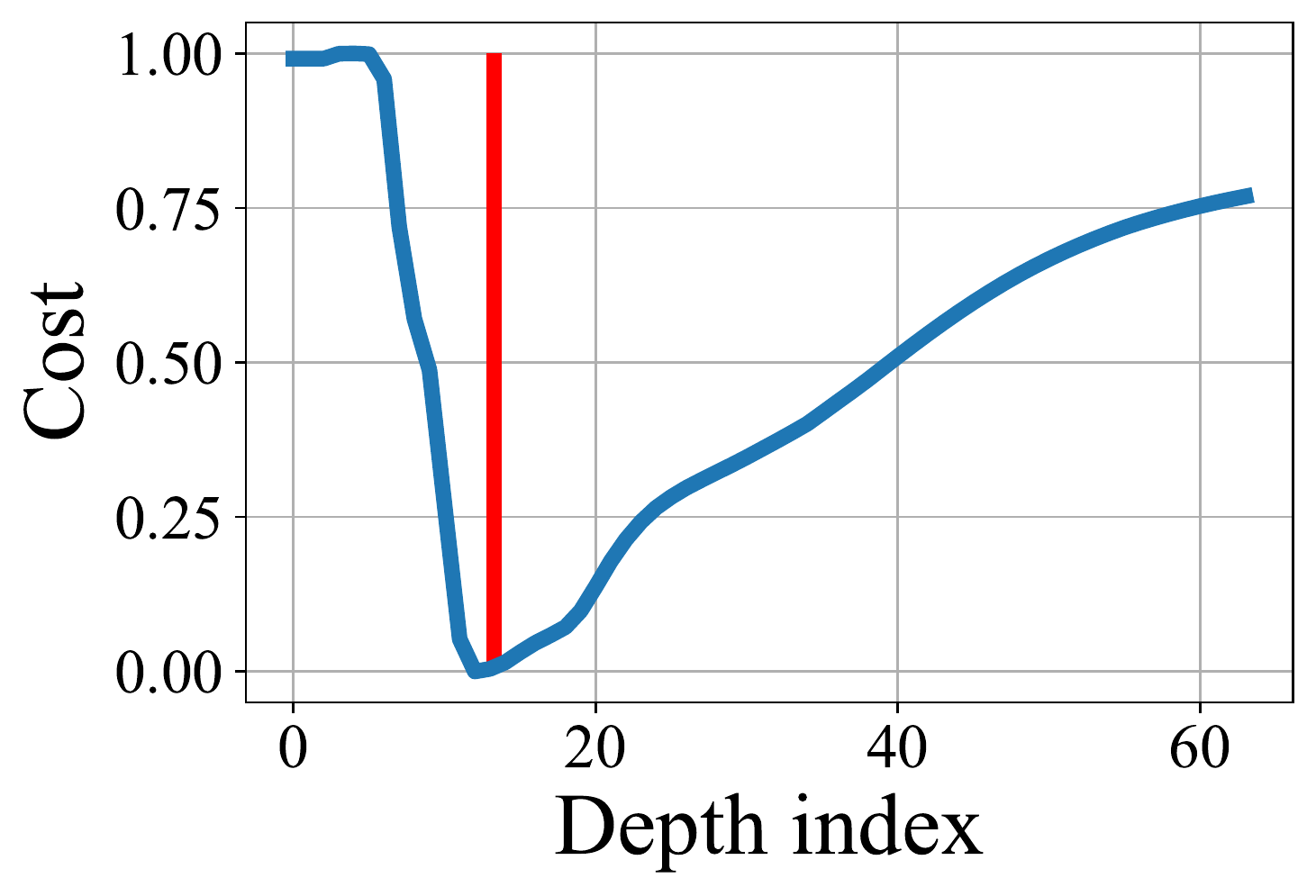}
    \subcaption{Outlier removal and normalization}
  \end{minipage}
  \captionsetup{format=plain}
  \caption{Cost plots (b) without and (c) with outlier removal and normalization at green dot in (a). Red lines indicate positions of ground-truth depth indices.}
  \label{fig:cost_plot}
\end{figure}

We also remove outliers by applying a nonlinear function $f(\cdot)$ that bounds the cost by 1 after computing Eq. (\ref{eq:cost_computation}).
We use a $\tanh$-like function as follows:
\begin{align}
f(x) &= \frac{e^{ax}-e^{-ax}}{e^{ax}+e^{-ax}},&\\
a &= \frac{1}{2C_{max}}\log\frac{1+f_1}{1-f_1},&
\end{align}
where $C_{max}$ is the upper bound of the cost.
$f(x)$ is converged to $f_1$ as $x$ approaches $C_{max}$.
We set $C_{max} = 0.3$ and $f_1=0.999$.
Finally, the cost $f(C(u,v,d))$ at each pixel is normalized in $[0,1]$.
As shown in Fig. \ref{fig:cost_plot}, this post-processing produces a sharp peak at the ground-truth depth.
However, this normalization includes the possibility that such sharp peaks also appear at texture-less pixels where defocus cues are not effective, thus have negative effects on training.
Nevertheless, we found that our network automatically learns effective regions and dramatically improves the accuracy of the estimated depth.
We describe the ablation study on this in Section \ref{sec:ablation_study}.

\subsection{Architecture and loss function}
As shown in Fig. \ref{fig:overview}, the cost volume and an additional defocus image, which helps the network to learn semantic cues \cite{Wang18}, are concatenated and passed through the network.
The input image is selected from the focal stack and we found that the selection of the input image does not affect the performance of the proposed method.
During the training of our model, we selected the image with the farthest focus distance.

The cost volume and input image are passed through the encoder, the architecture of which is the same as for MVDepthNet \cite{Wang18}.
The outputs of the decoder are refined cost volumes $C_{out}^s$ at different resolutions $s\in \{ 1/8, 1/4, 1/2, 1\}$.

\begin{figure}[tb]
\centering
  \captionsetup{format=hang}
  \begin{minipage}[t]{0.3\linewidth}
    \centering
    \includegraphics[width=1.0\textwidth]{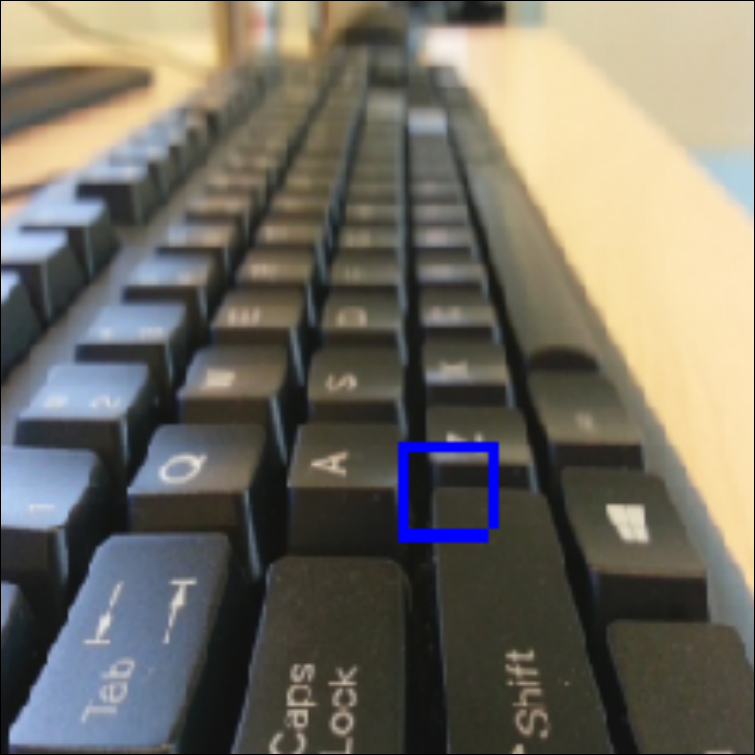}
    \subcaption{RGB}
  \end{minipage}
  \begin{minipage}[t]{0.3\linewidth}
    \centering
    \includegraphics[width=1.0\textwidth]{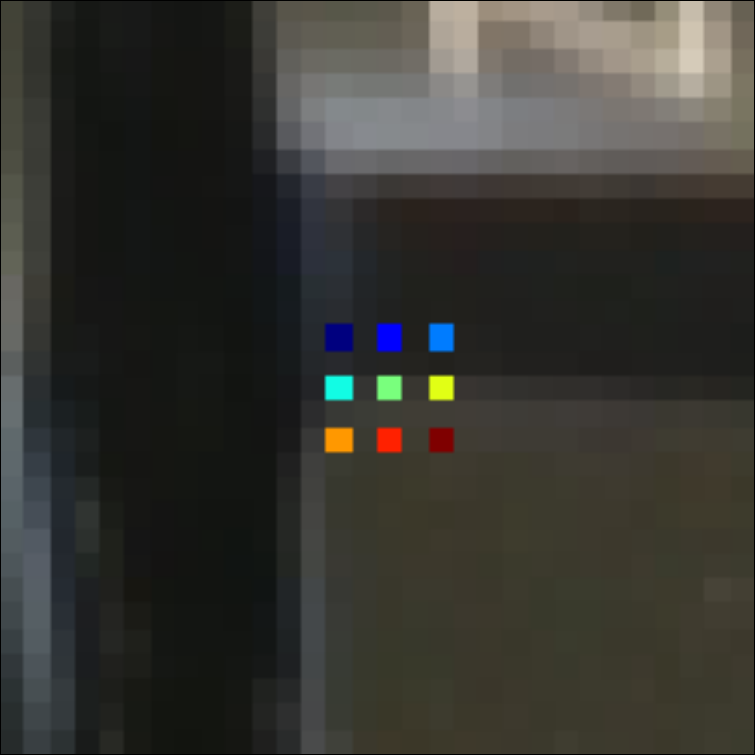}
    \subcaption{Initial offsets}
  \end{minipage}
  \begin{minipage}[t]{0.3\linewidth}
    \centering
    \includegraphics[width=1.0\textwidth]{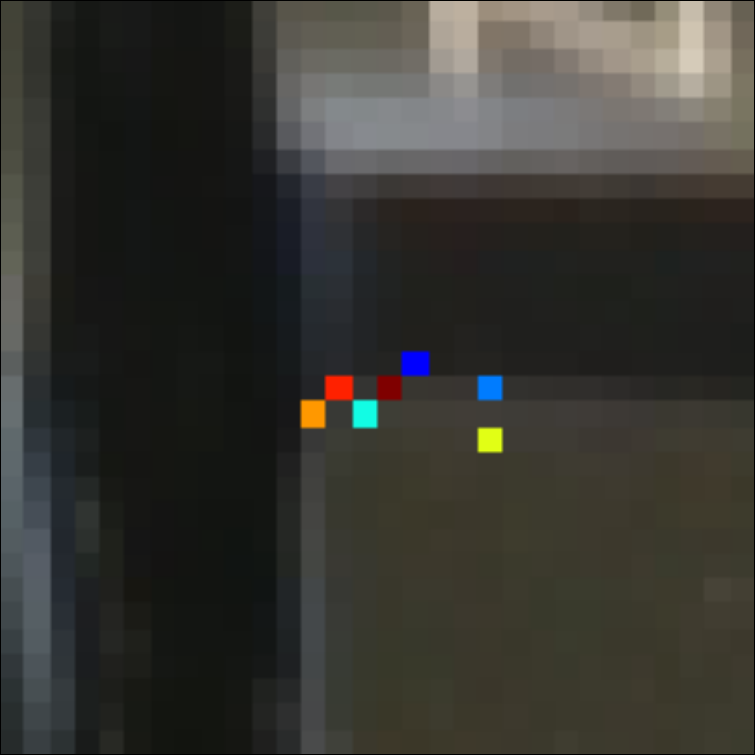}
    \subcaption{Learned offsets}
  \end{minipage}\\
  \begin{minipage}[t]{0.3\linewidth}
    \centering
    \includegraphics[width=1.0\textwidth]{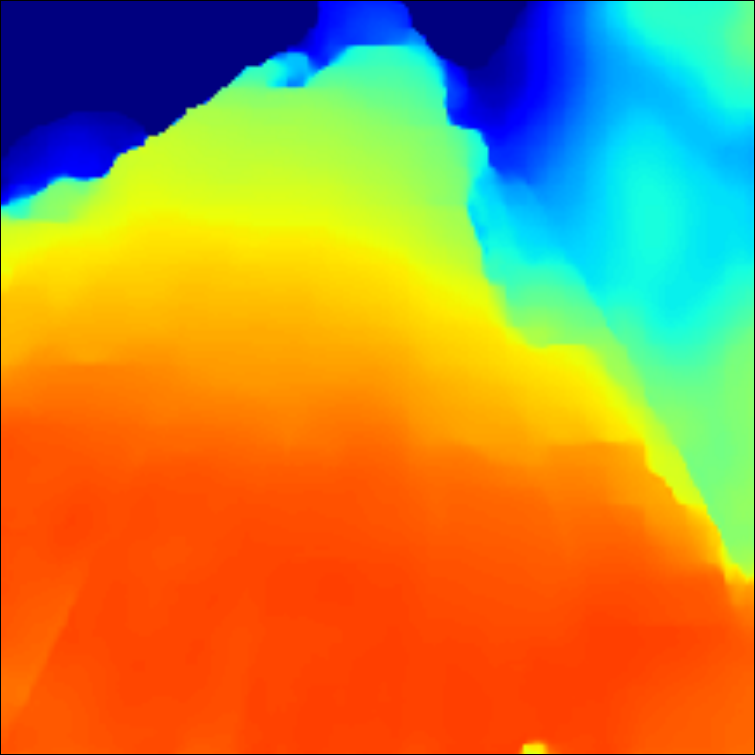}
    \subcaption{{\scriptsize w/o cost aggregation}}
  \end{minipage}
  \begin{minipage}[t]{0.3\linewidth}
    \centering
    \includegraphics[width=1.0\textwidth]{\figdir/keyboard_depth_pred.pdf}
    \subcaption{{\scriptsize w/ cost aggregation}}
  \end{minipage}
  \captionsetup{format=plain}
  \caption{Example of learned offsets for cost aggregation in blue boxed region in (a). (b) At beginning of training, cost is aggregated from nearby grid points. (c) After training, cost is adaptively aggregated by considering local structures. (e) This yields clear boundaries in estimated depth.}
  \label{fig:offset_plot}
\end{figure}

At each upsample block, we implement an adaptive cost aggregation module inspired by Wang \etal \cite{Wang21_CVPR} to aggregate neighboring information, and this enables depth estimation with clear boundaries by aggregating focus cues on edge pixels.
The cost aggregation module is given as
\begin{flalign}
&\tilde{C}_{out}^s(u,v,d_i) =& \nonumber \\
&\sum_{(u_j,v_j)\in \mathcal{N}(u,v)} w_j C_{out}^s(u_j + \Delta u_j, v_j + \Delta v_j, d_i),& \label{eq:cost_aggregation}
\end{flalign}
where the weight $w_j$ and offset $(\Delta u_j, \Delta v_j)$ are learnable parameters to aggregate neighboring information.
As shown in Fig. \ref{fig:overview}, our upsample block first upsamples the input cost volume by the scale factor of 2.
The feature map from the encoder is then concatenated to this upsampled cost volume.
From this volume, offsets and weights for adaptive cost aggregation are learned together with a refined cost volume.
The final cost volume is obtained by aggregating the neighboring costs following Eq. (\ref{eq:cost_aggregation}).
Figure \ref{fig:offset_plot} shows an example of the learned offsets and output depth with the cost aggregation module, which yields clear boundaries in the estimated depth.

The refined cost volume at each resolution is obtained through softmax layers.
Thus, the output depth at each resolution can be computed by applying a differentiable soft argmin operator \cite{Kendall17} as follows:
\begin{equation}
d_s(u,v) = \sum_{i} \tilde{C}_{out}^s(u,v,d_i) d_i.
\end{equation}

\paragraph{Training loss} The training loss is defined as the sum of L1 loss between the estimated depth maps $d_s$ and ground-truth depth maps $d_s^*$ at different resolutions as follows:
\begin{equation}
\mathcal{L} = \frac{1}{4}\sum_s \frac{1}{H_sW_s}\| d_s - d_s^* \|_1.
\end{equation}


\begin{table*}[tb]
  \caption{Camera settings of datasets}
  \label{tab:datasets}
  \centering
  \begin{adjustbox}{width=\textwidth}
    \begin{tabular}{cccccc}
      \hline
      Dataset & Size of focal stack & Focus distances $[\rm{m}]$ & Focal length $[\rm{m}]$ & f-number & $[\rm{m}/\rm{px}]$\\
      \hline
      DefocusNet \cite{Maximov20} & 5 & $\{0.1,0.15,0.3,0.7,1.5\}$ & $2.9\times 10^{-3}$ & $1$ & $1.2\times10^{-5}$  \\
      NYU Depth V2 \cite{Carvalho18} & 3 & $\{2,4,8\}$ & $15\times10^{-3}$& $2.8$ & $5.6\times10^{-6}$ \\
      \hline
    \end{tabular}
  \end{adjustbox}
\end{table*}

\section{Experiments}
We evaluated the proposed method for its camera-setting invariance and comparison it with the state-of-the-art learning-based DFF.
Our method can be applied to datasets with camera settings that differ from those of a training dataset.

\subsection{Implementation}
Our network was implemented in PyTorch. 
The training was done on a NVIDIA RTX 3090 GPU with 24-GB memory. 
The size of a minibatch was 8 for the training of our model.
We trained our network from scratch, and the optimizer was Adam \cite{Kingma15} with a learning rate of $1.0\times10^{-4}$.

During the cost volume construction, we uniformly sampled the depth between $0.1$ and $3$, and set the number of samples to $D=64$.

\begin{table}[tb]
  \caption{Ablation study for cost volume construction on DefocusNet dataset \cite{Maximov20}. Error metric is RMSE and errors were computed on datasets with different scales of data augmentation.}
  \label{tab:ablation_study_for_cv}
  \centering
    \begin{adjustbox}{max width=\linewidth}
    \begin{tabular}{ccccccc}
      \hline
       OR & Norm. & $\sigma = 1.0$ & $3.0$ & $5.0$ & $7.0$ & $9.0$\\
      \hline
      \checkmark & & 0.261 & 0.415 & 0.450 & 0.463 & 0.475  \\
       & \checkmark & {\bf 0.232} & {\bf 0.351} & 0.377 & 0.396 & 0.422 \\
      \checkmark & \checkmark & 0.239 & 0.363 & \bf{0.373} & \bf{0.380} & \bf{0.403} \\
      \hline
    \end{tabular}
    \end{adjustbox}
\end{table}

\begin{table}[tb]
  \caption{Experimental results on different focus distances at train and test time on DefocusNet dataset \cite{Maximov20}. Both methods were trained on focal stacks with focus distances $\{0.1,0.3,1.5\}$ then tested with focus distances $\{0.15,0.7\}$.}
  \label{tab:different_focus_distances}
  \centering
    \begin{tabular}{cccc}
      \hline
      Method & Train & Test & RMSE \\
      \hline
      DefocusNet \cite{Maximov20} & $\{0.1,0.3,1.5\}$ & $\{0.15,0.7\}$ & 0.299 \\
      Ours & $\{0.1,0.3,1.5\}$ & $\{0.15,0.7\}$ & \bf{0.242} \\
      \hline
    \end{tabular}
\end{table}

\begin{table*}[tb]
  \caption{Experimental results on blurred NYU Depth V2 dataset \cite{Carvalho18}. We computed errors on output depth and its rescaled version$^*$ because scales of output depths of current learning-based methods largely differ due to camera setting difference at training and test times. Scale-invariant errors in the depth space (sc-inv \cite{Eigen14}) and affine-invariant errors in inverse depth space (ssitrim \cite{Ranftl20}) were also computed for fair comparison.}
  \label{tab:nyu_depth_v2}
  \centering
    \begin{tabular}{ccccccc}
      \hline
      Method & Train dataset & MAE & RMSE & Abs Rel & sc-inv & ssitrim  \\
      \hline
      DDFF \cite{Hazirbas18} & DefocusNet & 0.719 & 0.773 & 0.793 & 0.199 & 0.318 \\
      AiFDepthNet \cite{Wang21} & DefocusNet & 0.425 & 0.491 & 0.412 & 0.319 & 0.509 \\
      DefocusNet \cite{Maximov20} & DefocusNet & 0.599 & 0.621 & 0.706 & 0.213 & {\bf 0.209}  \\ 
      \rowcolor[gray]{0.9}
      Ours & DefocusNet & {\bf 0.139} & {\bf 0.186} & {\bf 0.181} & {\bf 0.157} & {\bf 0.209} \\
      \hline \hline
      $^*$DDFF \cite{Hazirbas18} & DefocusNet & 0.138 & 0.313 & 0.165 & 0.199 & 0.318 \\
      $^*$AiFDepthNet \cite{Wang21} & DefocusNet & 0.239 & 0.312 & 0.276 & 0.319 & 0.509 \\
      $^*$DefocusNet \cite{Maximov20} & DefocusNet & 0.184 & 0.322 & 0.188 & 0.213 & {\bf 0.209} \\ 
      \rowcolor[gray]{0.9}
      $^*$Ours & DefocusNet & {\bf 0.097} & {\bf 0.141} & {\bf 0.126} & {\bf 0.157} & {\bf 0.209} \\
      \hline \hline
      DefocusNet \cite{Maximov20} & NYU Depth V2 & 0.016 & 0.029 & 0.018 & 0.030 & 0.033\\
      \rowcolor[gray]{0.9}
      Ours & NYU Depth V2 & 0.032 & 0.054 & 0.034 & 0.050 & 0.062\\
      \hline
      \multicolumn{6}{l}{$^*$Rescaled by median of ratios between output and ground-truth depths.}
    \end{tabular}
\end{table*}

\begin{figure*}[tb]
  \centering
  \includegraphics[width=1.0\textwidth]{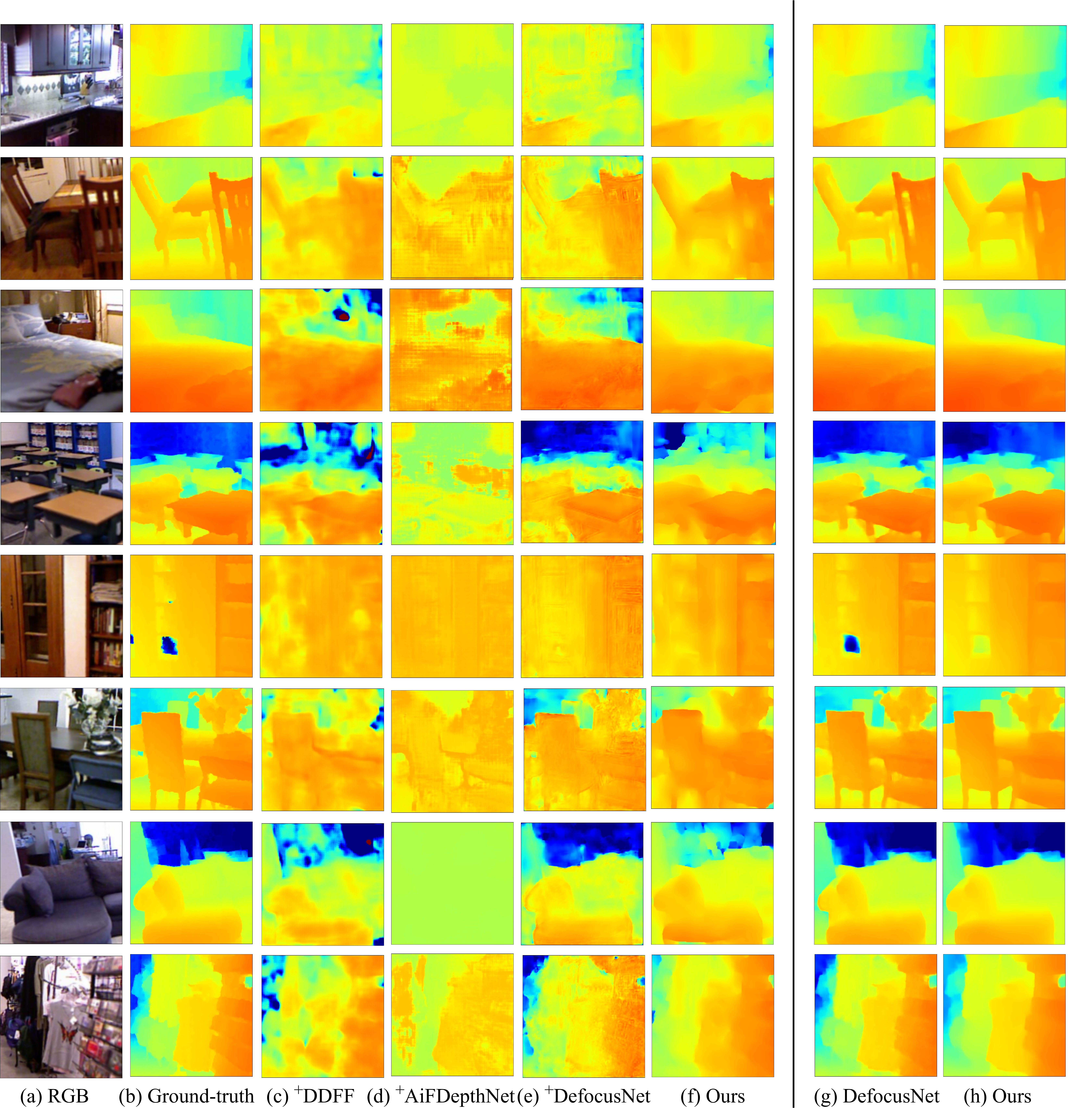}
  \caption{Qualitative comparison on NYU Depth V2 \cite{Carvalho18}. In (a)-(f), all models were trained on DefocusNet dataset \cite{Maximov20}. In (g) and (h), both methods were trained on NYU Depth V2 dataset. Superscript $^+$ means that affine-ambiguity is compensated by estimating scales and biases in least-squares manner between output and ground-truth.}
  \label{fig:nyu_depth_v2}
\end{figure*}

\begin{figure*}[tb]
\centering
  \captionsetup{format=hang}
  \begin{minipage}[t]{0.16\linewidth}
    \centering
    \includegraphics[width=1.0\textwidth]{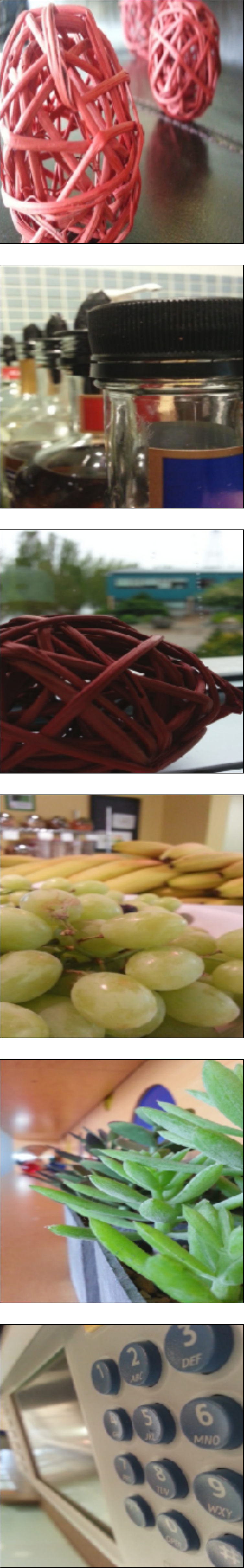}
    \subcaption{RGB}
  \end{minipage}
  \begin{minipage}[t]{0.16\linewidth}
    \centering
    \includegraphics[width=1.0\textwidth]{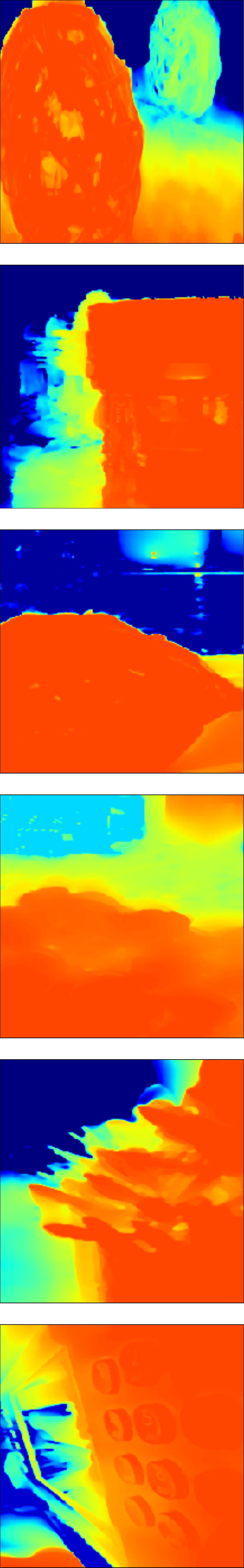}
    \subcaption{{\scriptsize Suwajanakorn \newline \etal \cite{Suwajanakorn15}}}
  \end{minipage}
  \begin{minipage}[t]{0.16\linewidth}
    \centering
    \includegraphics[width=1.0\textwidth]{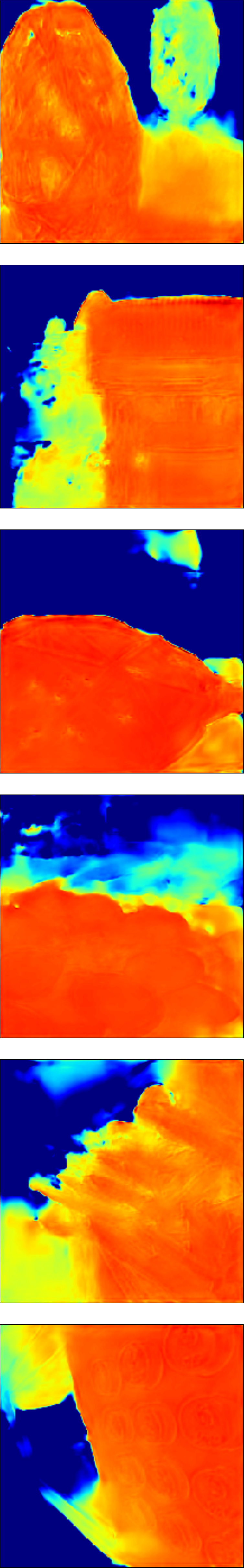}
    \subcaption{{\scriptsize $^*$DefocusNet \cite{Maximov20}}}
  \end{minipage}
  \begin{minipage}[t]{0.16\linewidth}
    \centering
    \includegraphics[width=1.0\textwidth]{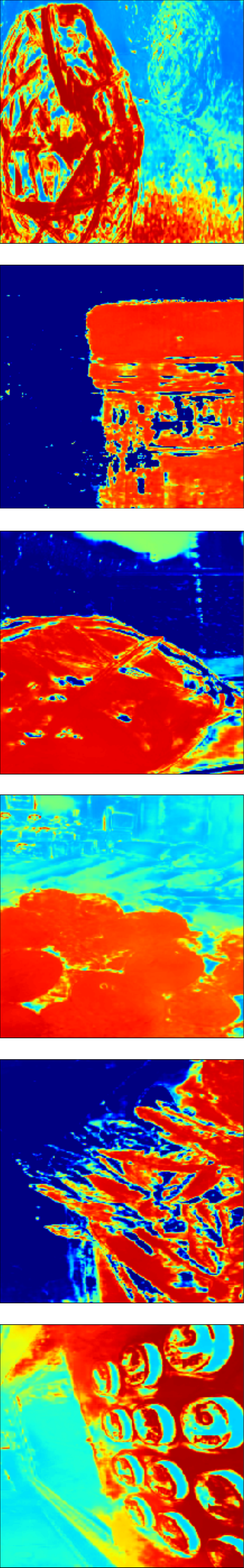}
    \subcaption{{\scriptsize $^*$AiFDepthNet \cite{Wang21}}}
  \end{minipage}
  \begin{minipage}[t]{0.16\linewidth}
    \centering
    \includegraphics[width=1.0\textwidth]{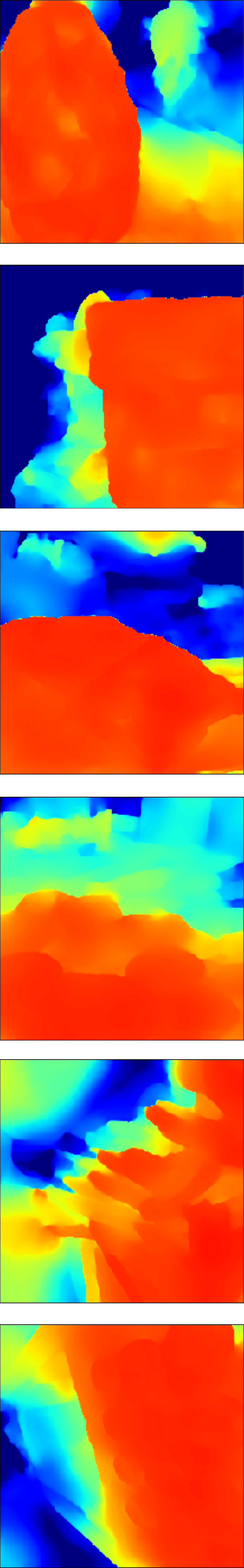}
    \subcaption{Ours}
  \end{minipage}
  \captionsetup{format=plain}
  \caption{Experimental results on Mobile Depth \cite{Suwajanakorn15}. Superscript $^*$ means that depth is rescaled by median of ratios between output and Suwajanakorn \etal \cite{Suwajanakorn15}.}
  \label{fig:mobile_depth}
\end{figure*}

\begin{figure}[tb]
  \centering
  \includegraphics[width=0.4\textwidth]{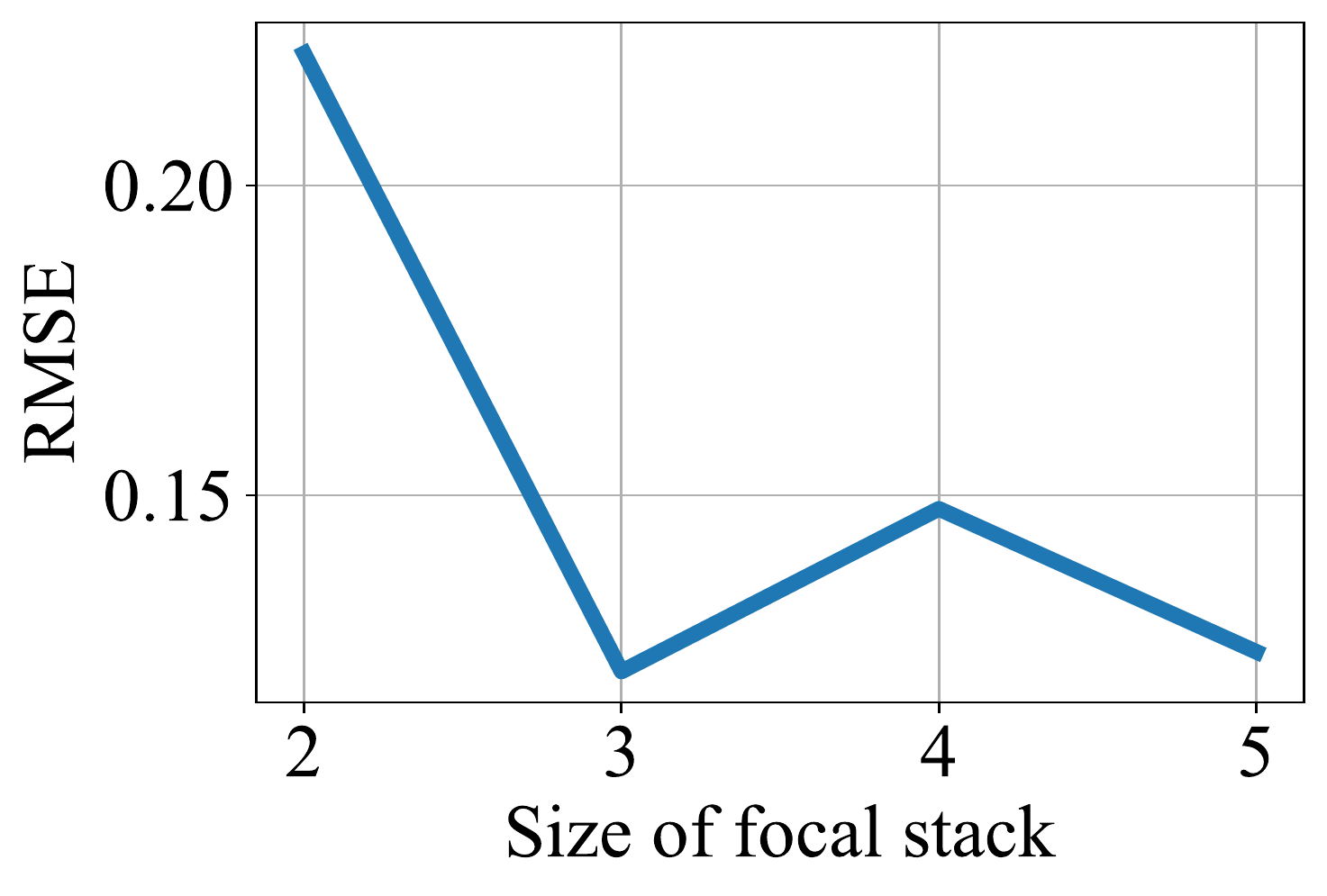}
  \caption{Ablation study of focal stack size on Mobile Depth \cite{Suwajanakorn15}. Horizontal and vertical axes represent RMSE and size of input focal stack.}
  \label{fig:ablation_focus_num}
\end{figure}

\begin{figure*}[tb]
\centering
  \includegraphics[scale=.35]{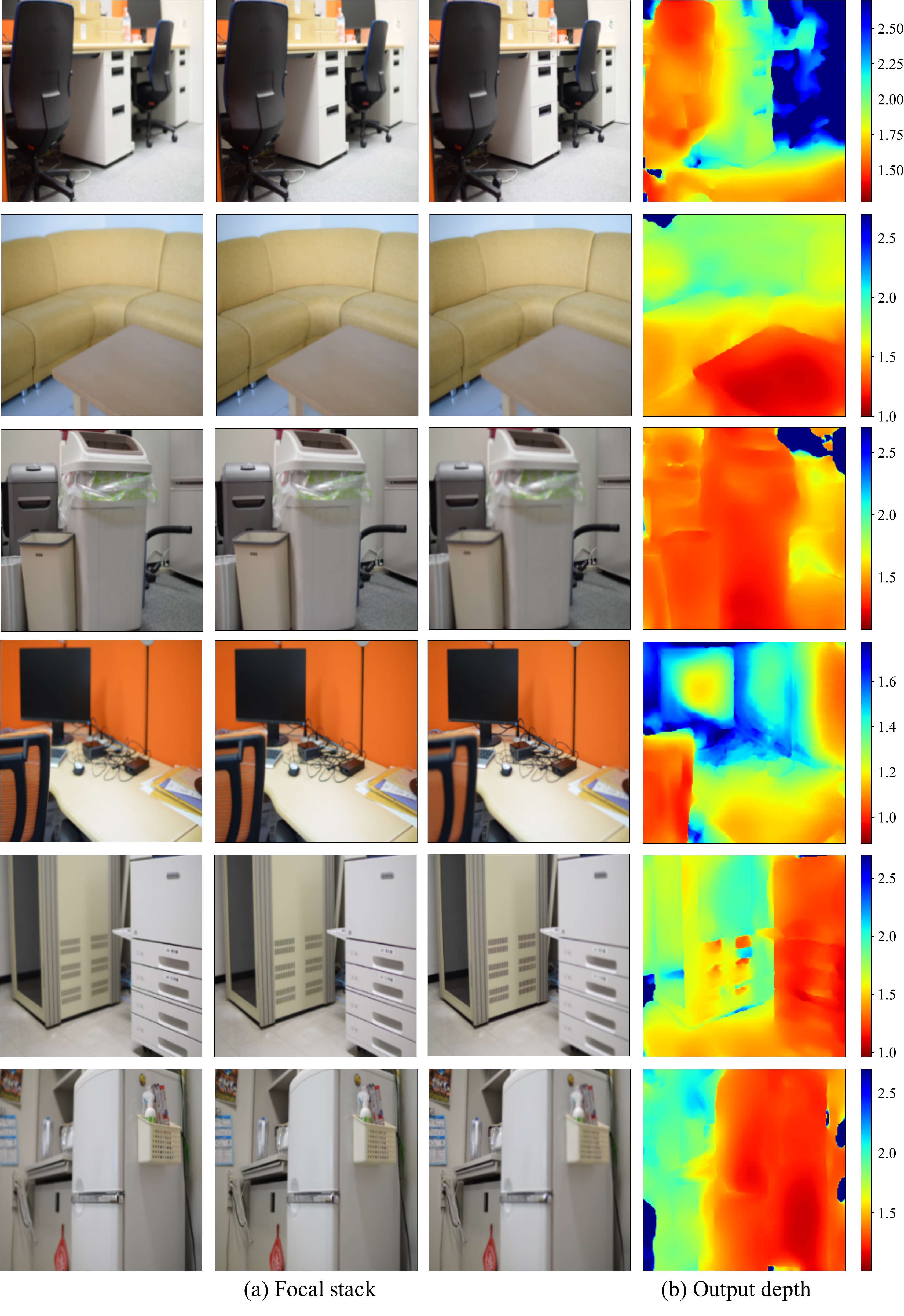}
  \caption{Experimental results on focal stacks captured with our camera. (a) Images in focal stack and (b) output depth of our method.}
  \label{fig:nikon_result}
\end{figure*}

\subsection{Dataset}
This section describes the datasets for training and evaluation.
We used three datasets with the meta data of full camera-settings.

\paragraph{DefocusNet dataset \cite{Maximov20}} This dataset consists of synthetic images, which were generated with physics-based rendering shaders on Blender.
The released subset of this dataset has 400 and 100 samples for training and evaluation, respectively.
The focal stack of each sample has five images with $256\times256$ resolution.
{\it Note that all models were trained only on this synthetic dataset unless otherwise noted.}

\paragraph{NYU Depth V2 \cite{Silberman12} synthetically blurred by \cite{Carvalho18}} Carvalho \etal \cite{Carvalho18} generated this dataset by adding synthetic blurs to the NYU Depth V2 dataset \cite{Silberman12} that consists of pairs of RGB and depth images.
The defocus model was based on Eq. (\ref{eq:defocus_model}) and takes into account object occlusions.
The official training and test splits of the NYU Depth V2 dataset are 795 and 654 samples.
We extracted $256\times256$ patches from the original $640\times480$ images and finally obtained 9540 and 7848 samples for training and evaluation.
As with \cite{Maximov20}, we rescaled the depth range from $[0,10]$ to $[0,3]$.
Table \ref{tab:datasets} lists the camera settings of the DefocusNet dataset \cite{Maximov20} and this NYU Depth V2 dataset \cite{Carvalho18}.

\paragraph{Mobile Depth \cite{Suwajanakorn15}}
This dataset consists of real focal stacks captured with a mobile phone camera.
The images in each focal stack were aligned and the authors estimated the camera parameters and depth (\ie, there are no actual ground-truth depth maps.).
This dataset contains several scenes; thus, we used this dataset only for evaluation.

\subsection{Data augmentation} \label{sec:data_augmentation}
In the DefocusNet dataset, defocus cues are effective only a short distance from a camera \cite{Maximov20}.
Therefore, we found that our cost volume learned on this dataset is effective only on small depth indices.
To enhance the scalability of our cost volume, we scaled the depth maps in the DefocusNet dataset by a scale factor of $\sigma \in \{1.0,1.5,2.0,\cdots,9.0\}$ when we trained our model on this dataset. 
We should also scale the camera parameters together with the depth map, \ie, if each data sample consists of $\{ \{I_{d_1},\cdots,I_{d_F}\}, \{d_1, \cdots, d_F \}, f, N, d^*, b  \}$, the scaled sample is $\{ \{I_{d_1},\cdots,I_{d_F} \}, \{\sigma d_1,\cdots, \sigma d_F \}, \sigma f, N, \sigma d^*, b/\sigma\}$.
Note that in both samples, the depth and camera parameters give the same amount of defocus blurs; thus the original focal stack can be used in the scaled sample.
This data augmentation is essential for applying our method to other datasets. 

\subsection{Ablation study} \label{sec:ablation_study}
Table \ref{tab:ablation_study_for_cv} lists the results from the ablation study on the cost volume construction.
We separately computed the RMSE on the DefocusNet dataset with a different scale factor of the data augmentation.
The experimental results demonstrate that normalization (Norm.) dramatically improved the accuracy of depth estimation.
Outlier removal (OR) also improved the accuracy, especially at a large depth scale, where the depth estimation will be more difficult than at a small depth scale, as mentioned in Section \ref{sec:data_augmentation}.

\subsection{Evaluation on different camera settings}
We then evaluated the performance of depth estimation with different camera settings at training and test times.
Table \ref{tab:different_focus_distances} lists the experimental results on the DefocusNet dataset.
DefocusNet \cite{Maximov20}, which is a state-of-the-art learning-based DFF method, was compared with our method.
We first decomposed each focal stack into two subsets, one with focus distances $\{0.1,0.3,1.5\}$ and the other with $\{0.15,0.7\}$.
Both methods were trained only on the subset with focus distances $\{0.1,0.3,1.5\}$ and evaluated on the other subset with different focus distances.
Our method outperformed DefocusNet, demonstrating the camera-setting invariance of our method.

We also evaluated the proposed method on the NYU Depth V2 dataset, which has different scene statistics and different camera settings from the DefocusNet dataset, as shown in Table \ref{tab:datasets}.
Table \ref{tab:nyu_depth_v2} and Fig. \ref{fig:nyu_depth_v2} show the experimental results when comparing the proposed method other with state-of-the-art learning-based methods, \ie, DDFF \cite{Hazirbas18}, AiFDepthNet \cite{Wang21}, and DefocusNet \cite{Maximov20}.
For AiFDepthNet, we used the authors' trained model, and the other methods were re-trained on the DefocusNet dataset.
The parameters of DDFF were initialized by VGG16 \cite{Simonyan15} as in the original paper \cite{Hazirbas18}.
For error metrics, we used MAE, RMSE, absolute relative L1 error (Abs Rel), scale-invariant error (sc-inv) \cite{Eigen14}, and affine- (scale- and shift-) invariant error in the inverse depth space denoted by ssitrim~\cite{Ranftl20}.

As shown in the upper part of Table \ref{tab:datasets}, our method outperformed the other methods trained on the DefocusNet dataset by large margins on most evaluation metrics, and is comparable to DefocusNet on the affine-invariant error metric in the inverse depth space (ssitrim).
This is because the camera settings of the DefocusNet and NYU Depth V2 datasets are different.
The other methods cannot handle this difference, and the estimated depths have ambiguity.

We also computed the errors on the depths rescaled by the median of the ratios between the output and the ground-truth depths followed by \cite{Maximov20} to compensate the scale-ambiguity.
The compensation has been done also on our results for fair comparison.
The errors are presented in the middle part of the table.
Our method also outperformed the other methods in this comparison.
In addition, our method without scaling (Ours) still outperformed the rescaled previous methods ($^*$) in most evaluation metrics.
Figure~\ref{fig:nyu_depth_v2} shows examples of the estimated depths.
In this figure, the affine-ambiguity of the other methods are compensated by estimating the scales and biases in a least-squares manner ($^+$).
Note that the output depths of our method were not rescaled, \ie, our method can estimate depths without any ambiguities.
In the bottom part of the table, we show the experimental results trained on the NYU Depth V2 dataset.
Although DefocusNet performed better than our method, the accuracy of both methods improved dramatically as shown in Figs. \ref{fig:nyu_depth_v2}(g) and (h), and DefocusNet is heavily affected by the difference of the camera settings in training and test datasets. 

Figure \ref{fig:mobile_depth} shows the experimental results on Mobile Depth with real focal stacks.
We set the size of an input focal stack to 3 except for AiFDepthNet \cite{Wang21}, which used from about 10 to 30 images for the size of a focal stack, and the model was trained on the synthetically blurred FlyingThings3D dataset \cite{Mayer16}.
The figure shows the qualitative comparison with the state-of-the-art learning-based methods, the output depths of which were rescaled by the median of the ratios between them and the outputs of Suwajanakorn \etal \cite{Suwajanakorn15} ($^*$) followed by \cite{Maximov20}.
Note that the output depths of our method were not rescaled.
The output depths of our method are qualitatively plausible and satisfy the defocus model under different camera settings.
Figure \ref{fig:ablation_focus_num} shows the quantitative errors between our method and Suwajanakorn \etal \cite{Suwajanakorn15} under different sizes of input focal stacks, demonstrating that a few images are enough to obtain effective results with our method.

Finally, we show an example of applying the proposed method to real focal stacks captured with our camera, Nikon D5300 with f-number of 1.8.
The focal stacks were captured with ``Focus Stacking Simple'' in digiCamControl \cite{digiCamControl}.
All parameters required for the cost volume computation were extracted from EXIF properties, and the focal stack size was 3.
Figure \ref{fig:nikon_result} shows the qualitative evaluation results.
The values of the estimated depth maps are in meters.
These results indicate the applicability of our method to real focal stacks.

\begin{table}[tb]
  \caption{Runtime comparison}
  \label{tab:runtime}
  \centering
    \begin{tabular}{cc|c}
      \hline
      \begin{tabular}{c}Cost volume\\construction\end{tabular} & Depth estimation & DefocusNet \cite{Maximov20} \\
      \hline
      4.278 sec. & 0.0252 sec. & 0.0285 sec. \\
      \hline
    \end{tabular}
\end{table}

\subsection{Computation time}
Table \ref{tab:runtime} shows the runtime comparison.
We measured the processing time for each test sample in the DefocusNet dataset \cite{Maximov20}.
The cost volume construction was done on AMD EPYC 7232P@3.1 GHz with 128GB RAM.
The number of the depth samples in the cost volume is 64 and the image resolution is $256 \times 256$.
Although the cost volume construction takes a few seconds, the costs at different depth slices in our cost volume can be computed in parallel to reduce the computation time.

\subsection{Limitations}
We finally discuss the limitations of the proposed methods due to the explicit lens defocus model.

\paragraph{Dynamic scenes and focus breathing}
Similar to AiFDepthNet \cite{Wang21}, our cost volume computation allows only static scenes.
Focus breathing also affects our method.
However, as mentioned in \cite{Wang21}, simple preprocessed alignment can solve this problem 
(In the experiments with real data (Fig. \ref{fig:mobile_depth}), we used aligned focal stacks).

\begin{figure}[tb]
\centering
  \captionsetup{format=hang}
  \begin{minipage}[t]{0.24\linewidth}
    \centering
    \includegraphics[width=0.9\textwidth]{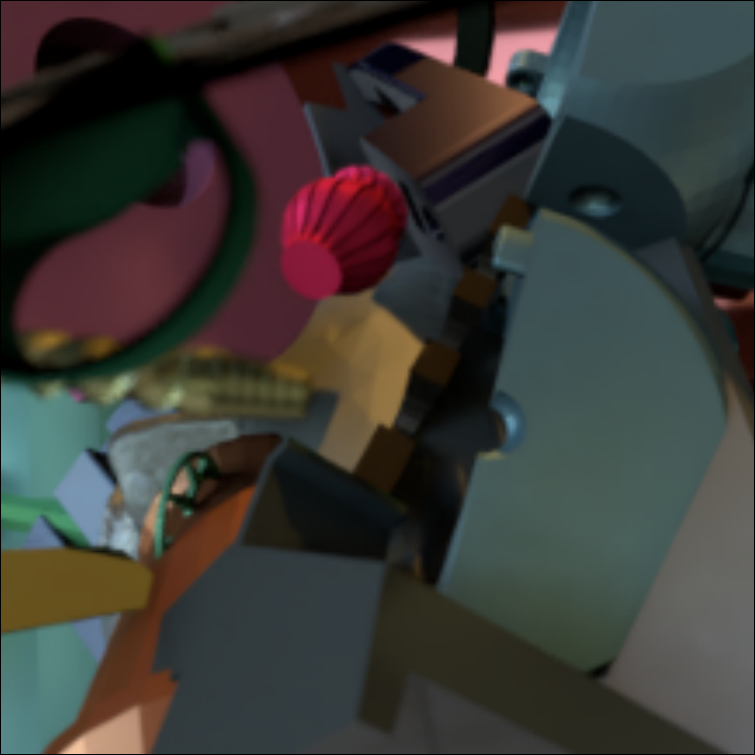}
    \subcaption{RGB}
  \end{minipage}
  \begin{minipage}[t]{0.24\linewidth}
    \centering
    \includegraphics[width=0.9\textwidth]{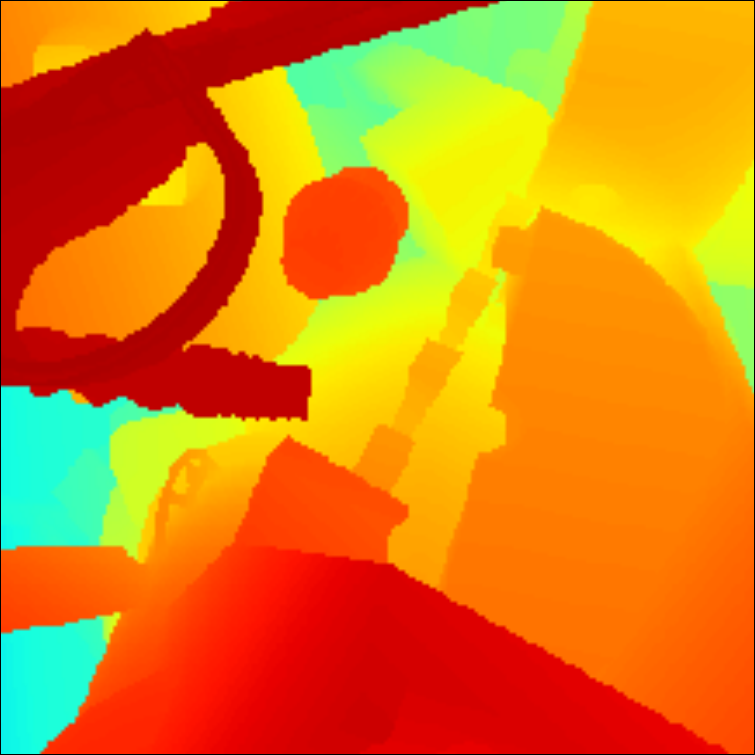}
    \subcaption{Ground-truth}
  \end{minipage}
  \begin{minipage}[t]{0.24\linewidth}
    \centering
    \includegraphics[width=0.9\textwidth]{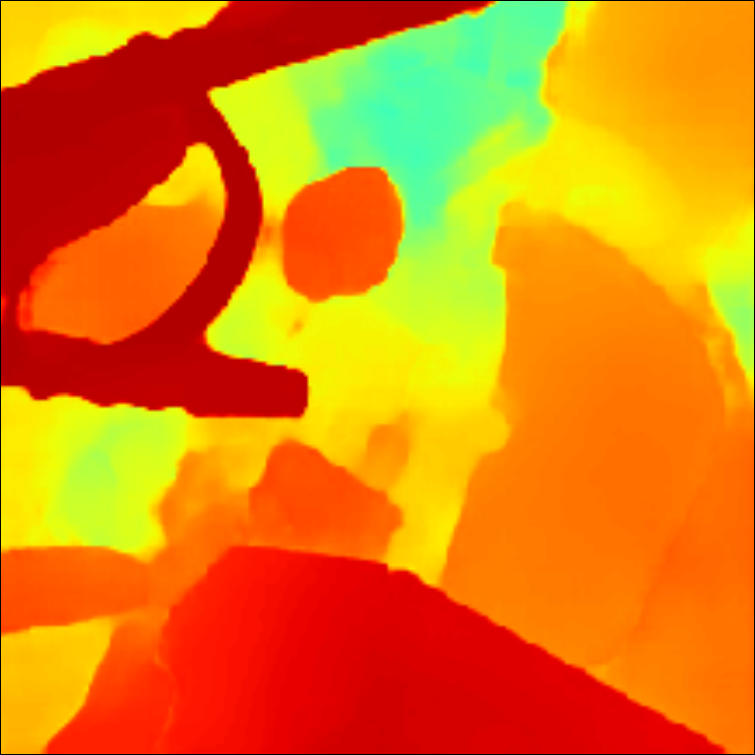}
    \subcaption{AiFDepthNet \cite{Wang21}}
  \end{minipage}
  \begin{minipage}[t]{0.24\linewidth}
    \centering
    \includegraphics[width=0.9\textwidth]{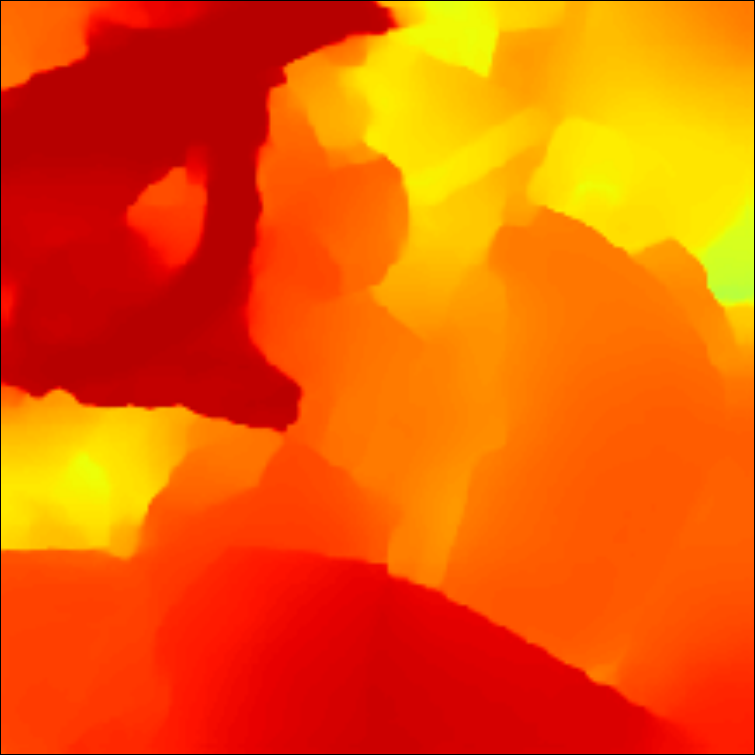}
    \subcaption{Ours}
  \end{minipage}
  \captionsetup{format=plain}
  \caption{Limitations of our method. (a) One of input images in focal stack, (b) Ground-truth depth, (c) output depth of AiFDepthNet \cite{Wang21}, and (d) that of our method. Defocus cues in the DefocusNet dataset are effective only at a short distance from the camera, and our method with explicit defocus model is more affected by this limitation.}
  \label{fig:limitation}
\end{figure}

\paragraph{Trade-off between defocus and semantic cues}
We finally discuss the trade-off between model- and learning-based approaches.
Table \ref{tab:defocus_net_result} and Fig. \ref{fig:limitation} show the results on the DefocusNet dataset.
The other learning-based methods outperformed our method.
This is because the defocus cues in the DefocusNet dataset are effective only at a short distance from a camera, as mentioned in Section \ref{sec:data_augmentation}.
The other learning-based methods handle this limitation through semantic cues.
Although our method also learns semantic cues, our method with the explicit lens defocus model is more affected by this limitation.
For future work, a network architecture should be designed to effectively learn defocus and semantic cues simultaneously.

\begin{table}[tb]
  \caption{Experimental results on DefocusNet dataset}
  \label{tab:defocus_net_result}
  \centering
    \begin{tabular}{cc}
      \hline
      Method & RMSE \\
      \hline
      AiFDepthNet \cite{Wang21} & 0.156 \\
      DefocusNet \cite{Maximov20} & 0.177 \\
      Ours & 0.239 \\
      \hline
    \end{tabular}
\end{table}

\section{Conclusion}
We proposed learning-based DFF with a lens defocus model.
We combined a learning framework and defocus model with the construction of a cost volume.
This method can absorb the difference in camera settings through the cost volume, which allows the method to estimate the scene depth from a focal stack with different camera settings at training and test times.
The experimental results indicate that our model trained only on a synthetic dataset can be applied to other datasets including real focal stacks with different camera settings.
This camera-setting invariance will enhance the applicability of learning-based DFF methods.

{\small
\bibliographystyle{ieee_fullname}
\bibliography{egbib}
}

\end{document}